\pdfoutput=1

\documentclass[11pt]{article}

\usepackage[preprint]{acl}

\usepackage{times}
\usepackage{latexsym}

\usepackage[T1]{fontenc}

\usepackage[utf8]{inputenc}

\usepackage{microtype}

\usepackage{inconsolata}

\usepackage{graphicx}

\usepackage{graphicx}
\usepackage{algorithm}
\usepackage{algpseudocode}
\usepackage{booktabs,chemformula}
\usepackage{amsfonts}
\usepackage[subrefformat=parens]{subcaption}
\usepackage{adjustbox}
\usepackage{cleveref}
\usepackage{makecell}
\usepackage{xcolor}
\usepackage{colortbl}
\usepackage{multirow}
\usepackage{mathabx}
\definecolor{bluee}{cmyk}{0.60, 0.19, 0.0, 0.08}
\definecolor{redd}{cmyk}{0.19, 0.60, 0.0, 0.08}
\usepackage{xcolor, tabularx, booktabs}
\usepackage{tcolorbox}
\usepackage{pifont}
\newcommand{\cmark}{\ding{51}}
\newcommand{\xmark}{\ding{55}}

\newcolumntype{u}{>{\columncolor{bluee!8}}c}
\newcolumntype{v}{>{\columncolor{bluee!24}}c}
\newcolumntype{d}{>{\columncolor{redd!8}}c}

\title{Evaluating Multimodal Generative AI with Korean Educational Standards}

\author{Sanghee Park$^\ast$\\
  NAVER Cloud AI\\
  \texttt{parksangheeeee@gmail.com} \\\And
  Geewook Kim$^\ast$$^\dagger$\\
  NAVER Cloud AI\\
  KAIST AI\\
  \texttt{gwkim.rsrch@gmail.com} \\}

\begin{document}
\maketitle

\begingroup
\renewcommand\thefootnote{}\footnote{$^\ast$ Sanghee Park and Geewook Kim contributed equally to this work and share first authorship.}
\renewcommand\thefootnote{}\footnote{$^\dagger$ Corresponding author.}%
\addtocounter{footnote}{-2}%
\endgroup

\begin{abstract}
This paper presents the Korean National Educational Test Benchmark (KoNET), a new benchmark designed to evaluate Multimodal Generative AI Systems using Korean national educational tests. KoNET comprises four exams: the Korean Elementary General Educational Development Test (KoEGED), Middle (KoMGED), High (KoHGED), and College Scholastic Ability Test (KoCSAT). These exams are renowned for their rigorous standards and diverse questions, facilitating a comprehensive analysis of AI performance across different educational levels. By focusing on Korean, KoNET provides insights into model performance in less-explored languages. We assess a range of models—open-source, open-access, and closed APIs—by examining difficulties, subject diversity, and human error rates. The code and dataset builder will be made fully open-sourced at \url{https://github.com/naver-ai/KoNET}.
\end{abstract}

\section{Introduction}
\label{sec:introduction}

\begin{figure}[t!]
  \centering
  \includegraphics[width=\linewidth]{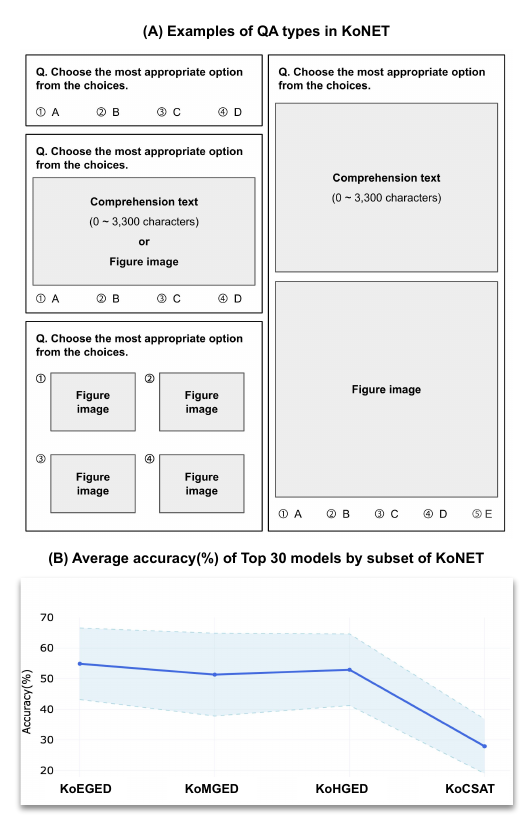}
\caption{\textbf{Examples and Performance Overview of KoNET.} (a) Illustration of mathematics problem examples, highlighting the increased complexity and difficulty as the educational level progresses. (b) Demonstration of how the accuracy of contemporary AI models decreases with more advanced curricula. A detailed analysis is provided in Section \ref{sec:exp_analysis}.}
  \label{fig:teaser}
\end{figure}

The advancement of Large Language Models (LLMs) has spurred the integration of sophisticated generative AI systems into various applications~\citep{openai2023gpt4}. Recent developments combining LLMs with computer vision have resulted in powerful Multimodal LLMs (MLLMs)~\citep{liu2023llava,liu2024llavanext,laurencon2024building,laurencon2024matters}. However, questions remain about the true intelligence of these systems, especially their ability to generalize across novel tasks similar to human cognition.

Current benchmarks predominantly focus on English, overlooking the linguistic diversity worldwide and offering limited insights into low-resource languages like Korean. Moreover, many benchmarks do not compare AI performance to that of humans, making it difficult to precisely measure AI proficiency. Some benchmarks are also less connected to real-world application scenarios, hindering the applicability of MLLMs.

To address these challenges, we introduce KoNET, a benchmark dataset leveraging four key Korean educational tests (refer to Figure \ref{fig:teaser}). Each exam—KoEGED, KoMGED, KoHGED, and KoCSAT—provides detailed analyses of question difficulty, enabling nuanced evaluation of AI capabilities. Notably, KoCSAT includes data on the percentage of incorrect responses per item among examinees (human error rate), facilitating thorough comparisons of model behaviors with human performance. This benchmark allows for direct comparisons to human performance and underscores essential competencies crucial for AI-driven educational technologies, offering potential real-world applicability in the AI tutoring market.

Our key contributions include:
\begin{enumerate}
    \item The introduction of KoNET, a comprehensive benchmark for evaluating Multimodal Generative AI Systems via Korean educational tests.
    \item A thorough evaluation of various open-source, open-access, and closed API models.
    \item Insights through multiple analytical frameworks, examining the relationship between human and model error rates.
\end{enumerate}

\begin{table}[t]
\begin{adjustbox}{width=\linewidth}
  \centering
  \begin{tabular}{lcccc}
    \toprule
    \textbf{Statistic} & \textbf{KoEGED} & \textbf{KoMGED} & \textbf{KoHGED} & \textbf{KoCSAT}\\
    \hline
    Images & 400 & 540 & 540 & 897 \\
    Questions & 400 & 540 & 540 & 897 \\
    \: $^{\ast}$K-QA & 62 (15.5\%) & 65 (12.0\%) & 62 (11.5\%) & 57 (6.4\%) \\
    \: $^{\dagger}$TC-QA & 123 (30.8\%) & 249 (46.1\%) & 284 (52.6\%) & 388 (43.3\%) \\
    \: $^{\ddagger}$MC-QA & 215 (53.8\%) & 226 (41.9\%)	& 194 (35.9\%) & 452 (50.3\%) \\
    \hline
    Subjects & 10 & 11 & 11 & 41 \\
    \hline
    Choices & 4 (100.0\%) & 4 (100.0\%) & 4 (100.0\%) & 5 (98.8\%) \\
    \hline
    Avg word & 29.9 & 42.7 & 48.0 & 113.0 \\
    Max word & 106 & 362 & 410 & 786 \\
    Avg Char & 113.0 & 167.2 & 193.6 & 475.9 \\
    Max Char & 417 & 1,408 & 1,678 & 3,300 \\
    \hline
    \#choice                & 4 & 4 & 4 & 5 \\
    \bottomrule
  \end{tabular}
\end{adjustbox}
  \caption{\label{tab:key_stats}
    \textbf{Key statistics of the KoNET benchmark.} $^{\ast}$K-QA: Knowledge QA, $^{\dagger}$TC-QA: Text Comprehension QA, and $^{\ddagger}$MC-QA: Mutimodal Comprehension QA.
  }
\end{table}

\begin{table}[t]
\begin{adjustbox}{width=\linewidth}
  \centering
  \begin{tabular}{cccccccc}
    \toprule
    \textbf{Bench} & \textbf{Lang.} & \textbf{\#Q} & \textbf{\#I} & \textbf{\#choice}  & $^{\ast}$\textbf{D} & $^{\dagger}$\textbf{H} \\
    \hline
    AI2D & En & 3,088 & 3,088 & $=4$ (100.0\%) & \xmark & \xmark \\
    ScienceQA & En & 4,240 & 2,017 & $\leq5$ (100.0\%) & \xmark & \xmark \\
    MMMU & En & 900 & 1,900 & $\leq9$ (94.1\%) & \cmark & \xmark \\
    Mathvista & En & 1,000 & 1,000 & $\leq8$ (53.4\%) & \cmark & \xmark \\
    \hline
    \rowcolor{bluee!8} \textbf{KoNET (ours)}    & Ko & 2,377 & 2,377 & $\leq5$ (99.5\%) & \cmark & \cmark \\
    \bottomrule
  \end{tabular}
\end{adjustbox}
\caption{\label{tab:compare_mcqa} \textbf{Comparison of Multiple-Choice QA Public Benchmarks.} $^{\ast}$D indicates that difficulty levels are provided for each question, and $^{\dagger}$H denotes that human error rate data is available for certain items.}
\end{table}

\section{Related Work}
\label{sec:related_work}

\paragraph{Text Benchmarks.}

MMLU~\citep{hendrycks2021measuring} assesses general language proficiency, while GSM8K~\citep{cobbe2021gsm8k}, CS-Bench~\citep{song2024cs}, and SciBench~\citep{wang2024scibench} focus on math, computer science, and science skills. These offer a focused evaluation of AI capabilities within educational contexts.

\paragraph{Multimodal Benchmarks.}

SEEDBench~\citep{li2023seed} and MMStar~\citep{chen2024we} provide general multimodal evaluations. Notably, there are educationally focused benchmarks such as ScienceQA~\citep{lu2022learn} and MathVista~\citep{lu2024mathvista}, which assess AI’s ability with scientific and mathematical content. Further, MMMU~\citep{yue2023mmmu} provides diverse subject evaluations, including Art and Medicine, while AI2D~\citep{Kembhavi2016ADI} examines diagram interpretation in grade school science.

\paragraph{Korean Benchmarks.}
Korean benchmarks are limited, but efforts like K-MMLU~\citep{son2024kmmlumeasuringmassivemultitask} and Ko-H5~\citep{park-etal-2024-open} have emerged. In multimodal contexts, KVQA~\citep{Kim_Lim2019} and CVQA~\citep{romero2024cvqa} focus on VQA and cultural understanding.
Despite the advances, there is a notable absence of Korean educational benchmarks, particularly in the multimodal domain. No existing frameworks comprehensively evaluate AI's educational performance across various school subjects within a Korean context.

\section{Proposed Benchmark: KoNET}
\label{sec:benchmark_construction}

To offer a robust evaluation framework that facilitates comprehensive comparisons with human educational levels, we converts questions from Korea's national educational tests into a multimodal VQA format. Table~\ref{tab:key_stats} presents key statistics of KoNET, while Table~\ref{tab:compare_mcqa} shows its main contributions.

\subsection{Education System and Qualification Exams in Korea}

Education is core to societal progress in Korea, with a structured system consisting of 6 years in elementary, 3 in middle, 3 in high school, and 4 in university or 2-3 in junior college~\citep{koreaneducentreinukEducationKorea}.

The \textbf{General Educational Development (GED)} exams assess basic academic knowledge for individuals who have not completed formal schooling, granting qualifications equivalent to traditional graduation upon passing. The \textbf{College Scholastic Ability Test (CSAT)}, also known as ``Suneung,'' is instrumental for college admissions and is recognized for its difficulty and ability to distinguish academic excellence.

\subsection{Construction of KoNET}

KoNET is constructed by parsing publicly available official PDFs from the Korea Institute of Curriculum and Evaluation\footnote{\url{https://www.kice.re.kr}}. The GED tests include all questions from the first and second sessions of 2023, with each exam comprising 20 or 25 multiple-choice questions per subject, with four options provided for each question. The CSAT incorporates questions from various subjects conducted in 2023, with a range of 20 to 45 questions each. While most are multiple-choice, some subjects have subjective questions. For the CSAT, human error rates are available for a selective subset of 327 questions. This subset reflects the challenges and complexities of these questions, as human error rate data is disclosed primarily for items with higher difficulty levels. Each data sample in KoNET is represented by a single image. More details are in Appendix~\ref{appendix:examples_knet}.

\section{Experiment and Analysis}
\label{sec:exp_analysis}

\subsection{Setup}

To thoroughly test contemporary models, we use 18 open-source LLMs, 20 open-source MLLMs, 4 closed-source LLMs, and 4 closed-source MLLMs, covering a range of sizes and complexities.

\paragraph{Response Generation.}
We employ the Chain-of-Thought (CoT)~\citep{NEURIPS2022_9d560961} as some KoNET problems requires complex reasoning. We use the OCR API\footnote{\url{https://www.ncloud.com/product/aiService/ocr}}, specialized for Korean, to translate image content for LLM models lacking vision capabilities. MLLMs use OCR as supplementary information. The ablations on CoT prompting and OCR are in Section~\ref{sec:exp_analysis}. The CoT prompts used in this study are in Appendix~\ref{appendix:prompts}. In this study, we ensured a consistent evaluation environment for LLMs and MLLMs across multiple benchmarks, including KoNET, MMMU, and MathVista, using a unified prompt structure and input format. Recent multimodal benchmarks like MMMU-Pro ~\citep{yue2024mmmu-pro} and EXAMS-V ~\citep{das2024exams} embed all necessary information within images, requiring MLLMs to extract and interpret content directly. KoNET follows this approach, incorporating both questions and answer choices into images, eliminating the need for explicit {question} and {option} placeholders (Figure \ref{fig:prompt}). LLMs do not receive direct textual inputs but can infer information via OCR-extracted text. Furthermore, KoNET includes problems where answer choices are images rather than text, requiring MLLMs to rely on visual reasoning. This design enables a more realistic assessment of multimodal comprehension and reasoning abilities.

\paragraph{Evaluation.}
We utilize the LLM-as-a-Judge approach~\citep{zheng2023judging} with GPT-4o~\cite{openai2023gpt4} to verify correctness. This method eliminates the need for manually parsing each model output, thereby minimizing potential errors.

\begin{table*}[t!]
\begin{adjustbox}{width=\linewidth}
  \centering
  \begin{tabular}{lcccccuuuuv}
    \toprule
     \multirow{2}{*}{\textbf{Model}} & \multirow{2}{*}{\textbf{Size (B)}} & \multicolumn{4}{c}{\textit{Previous Benchmarks}} & \multicolumn{4}{c}{\textit{\textbf{Proposed KoNET Benchmarks}}} &  \\
     \cline{3-6}\cline{7-10}
    & & \textbf{Mathvista} & \textbf{ScienceQA} & \textbf{AI2D} & \textbf{MMMU} & \textbf{KoEGED} & \textbf{KoMGED} & \textbf{KoHGED} & \textbf{KoCSAT} & \multirow{-2}{*}{\cellcolor{bluee!24}\textbf{KoNET}} \\
     \hline
     Open Source LLM \\
     \hline
     Qwen2-0.5B-Instruct~\citep{yang2024qwen2technicalreport}                              & 0.5  &	4.9 &	29.8 &	20.2 &	 4.5 &	17.8 &	19.6 &	16.7 &	12.8 &	16.0 \\
     Qwen2-1.5B-Instruct~\citep{yang2024qwen2technicalreport}                              & 1.5  &	2.8 &	32.6 &	19.6 &	 6.1 &	25.8 &	20.6 &	22.0 &	14.3 &	19.2 \\
     gemma-2-2b-it~\cite{gemmateam2024gemmaopenmodelsbased}                                & 2.0  &	1.0 &	30.0 &	24.7 &	 9.8 &	30.0 &	30.7 &	32.4 &	16.5 &	25.3 \\
     Phi-3-mini-4k-instruct~\cite{abdin2024phi}                     & 3.8  & 5.1 &	31.4 &	26.1 &	14.1 &	37.0 &	37.0 &	37.4 &	18.1 &	29.5 \\
     Phi-3.5-mini-instruct~\cite{abdin2024phi}                      & 3.8  &	5.5 &	34.9 &	26.8 &	10.9 &	29.0 &	28.0 &	23.5 &	14.6 &	21.8 \\
     Yi-1.5-6B-Chat~\cite{young2024yi}                              & 6.0  &	5.2 &	33.8 &	25.6 &	14.2 &	39.2 &	36.7 &	36.1 &	19.7 &	30.2 \\
     Mistral-7B-Instruct-v0.3\cite{jiang2023mistral}                & 7.0  &	7.6 &	36.7 &	34.2 &	20.5 &	36.5 &	29.4 &	34.4 &	16.5 &	26.5 \\
     Qwen2-7B-Instruct~\citep{yang2024qwen2technicalreport}                                & 7.0  &	6.4 &	35.4 &	33.2 &	23.3 &	54.0 &	53.1 &	50.7 &	20.3 &	39.6 \\
     EXAONE-3.0-7.8B-Instruct~\cite{exaone-3.0-7.8B-instruct}       & 7.8  & 7.1 &	39.3 &	34.1 &	21.9 &	64.5 &	59.1 &	56.9 &	24.2 &	45.5 \\
     Meta-Llama-3-8B-Instruct\cite{dubey2024llama3herdmodels}       & 8.0  &	6.0 &	37.3 &	39.2 &	22.3 &	46.5 &	46.9 &	43.3 &	20.5 &	35.5 \\
     Meta-Llama-3.1-8B-Instruct\cite{llama31}                       & 8.0  &	5.3 &	38.2 &	36.7 &	19.7 &	42.5 &	41.9 &	40.6 &	18.4 &	32.3 \\
     Yi-1.5-9B-Chat~\cite{young2024yi}                              & 9.0  &	8.2 &	37.5 &	38.6 &	20.7 &	47.0 &	43.7 &	45.0 &	22.5 &	36.0 \\
     gemma-2-9b-it~\cite{gemmateam2024gemmaopenmodelsbased}                                & 9.0  &	6.7 &	41.7 &	41.8 &	20.0 &	63.0 &	61.3 &	59.3 &	29.8 &	48.5 \\
     Phi-3-medium-4k-instruct~\cite{abdin2024phi}                   & 14.0 & 12.6 &	48.7 &	41.6 &	17.3 &	34.8 &	34.8 &	32.0 &	17.7 &	27.4 \\
     gemma-2-27b-it~\cite{gemmateam2024gemmaopenmodelsbased}                               & 27.0 &	18.8 &	49.6 &	47.3 &	24.6 &	74.5 &	69.6 &	68.5 &	33.9 &	55.9 \\
     Yi-1.5-34B-Chat~\cite{young2024yi}                             & 34.0 &	18.9 &	61.5 &	44.2 &	25.1 &	64.0 &	57.4 &	55.4 &	25.8 &	45.4 \\
     Meta-Llama-3.1-70B-Instruct\cite{llama31}                      & 70.0 & 20.3 &	67.5 &	49.5 &	31.5 &	63.2 &	65.6 &	62.6 &	31.2 &	50.8 \\
     Qwen2-72B-Instruct~\citep{yang2024qwen2technicalreport}                               & 72.0 & 21.7 &	69.1 &	49.4 &	32.3 &	76.0 &	74.1 &	71.9 &	36.0 &	58.7 \\
     \hline
     Open Source VLM \\
     \hline
     InternVL2-1B~\cite{chen2024far}                    &   1.0 & 33.5 &	59.6 &	65.2 &	35.0 &	 0.8 &	 0.4 &	 0.9 &	 0.4 &	 0.6 \\
     InternVL2-2B~\cite{chen2024far}                    &   2.0 & 35.4 &	62.0 &	74.0 &	35.7 &	 2.2 &	 2.0 &	 3.3 &	 1.7 &	 2.2 \\
     Qwen2-VL-2B-Instruct~\cite{Qwen2VL}                &   2.0 & 42.9 &	65.4 &	76.5 &	40.2 &	13.2 &	13.0 &	12.2 &	 8.4 &	11.0 \\
     paligemma-3b-mix-448~\cite{beyer2024paligemma}     &   3.0 & 29.1 &	65.3 &	69.8 &	33.4 &	 8.2 &	 8.7 &	 8.7 &	 4.9 &	 7.1 \\
     InternVL2-4B~\cite{chen2024far}                    &   4.0 & 57.0 &	71.5 &	78.7 &	46.5 &	 1.5 &	 2.0 &	 1.7 &	 0.9 &	 1.4 \\
     Phi-3.5-vision-instruct~\cite{abdin2024phi}        &   4.2 & 44.8 &	68.6 &	77.8 &	39.3 &	15.0 &	17.0 &	13.1 &	 4.6 &	10.9 \\
     Qwen2-VL-7B-Instruct~\cite{Qwen2VL}                &   7.0 & 53.2 &	66.7 &	71.5 &	59.1 &	49.5 &	46.9 &	42.0 &	16.9 &	34.3 \\
     llava-1.5-7b-hf~\cite{liu2024improved}             &   7.0 & 30.9 &	67.3 &	53.0 &	30.8 &	 3.2 &	 4.6 &	 4.8 &	 3.2 &	 3.9 \\
     llava-v1.6-vicuna-7b-hf~\cite{liu2024llavanext}    &   7.0 & 35.2 &	71.7 &	53.9 &	34.0 &	 3.0 &	 2.8 &	 1.9 &	 1.6 &	 2.1 \\
     InternVL2-8B~\cite{chen2024far}                    &   8.0 & 58.2 &	61.9 &	65.9 &	53.3 &	12.2 &	11.7 &	 8.0 &	 4.0 &	 7.9 \\
     llama3-llava-next-8b-hf~\cite{liu2024llavanext}    &   8.0 & 37.1 &	70.5 &	55.8 &	35.1 &	10.2 &	 7.8 &	 7.2 &	 2.6 &	 6.0 \\
     llava-1.5-13b-hf~\cite{liu2024improved}            &  13.0 & 26.6 &	49.3 &	57.6 &	37.5 &	11.8 &	 8.1 &	 7.4 &	 4.6 &	 7.1 \\
     llava-v1.6-vicuna-13b-hf~\cite{liu2024llavanext}   &  13.0 & 37.0 &	71.5 &	60.3 &	34.9 &	 5.0 &	 5.0 &	 7.2 &	 6.9 &	 6.3 \\
     cogvlm2-llama3-chat-19B~\cite{hong2024cogvlm2}     &  19.0 & 40.0 &	59.3 &	74.7 &	43.5 &	 5.8 &	 6.7 &	 4.6 &	 6.1 &	 5.9 \\
     InternVL2-26B~\cite{chen2024far}                   &  26.0 & 59.5 &	60.3 &	84.4 &	46.6 &	 8.8 &	 6.5 &	 7.2 &	 1.3 &	 5.0 \\
     llava-v1.6-34b-hf~\cite{liu2024llavanext}          &  34.0 & 44.6 &	63.6 &	83.6 &	50.7 &	25.0 &	 0.0 &	50.0 &	 0.0 &	15.0 \\
     InternVL2-40B~\cite{chen2024far}                   &  40.0 & 58.3 &	70.5 &	87.7 &	51.5 &	49.3 &	 0.0 &	36.8 &	11.9 &	20.8 \\
     llava-next-72b-hf~\cite{liu2024llavanext}          &  72.0 & 51.9 &	79.4 &	77.1 &	44.9 &	49.0 &	45.0 &	39.4 &	10.6 &	30.7 \\
     InternVL2-Llama3-76B~\cite{chen2024far}            &  76.0 & 64.1 &	81.7 &	87.0 &	55.1 &	10.9 &	 7.3 &	11.1 &	 4.3 &	 7.5 \\
     llava-next-110b-hf~\cite{liu2024llavanext}         & 110.0 & 55.1 &	85.4 &	83.1 &	48.7 &	19.8 &	23.0 &	20.9 &	12.0 &	17.6 \\
     \hline
     Closed Source LLM \\
     \hline
     gemini-1.5-pro(2024.05)\cite{google2024gemini15}       & N/A &	19.1 &	68.3 &	53.9 &	32.7 &	80.0 &	81.7 &	81.9 &	44.0 &	66.4 \\
     HyperCLOVA-X(2024.09)\cite{yoo2024hyperclova}          & N/A &	20.9 &	83.8  &	50.7 &	29.1 &	82.0 &	84.6 &	85.1 &	51.2 &	70.9 \\
     claude-3-5-sonnet-20240620\cite{anthropic2024claude35} & N/A &	27.6 &	80.0 &	61.5 &	54.2 &	86.5 &	86.3 &	86.1 &	60.5 &	76.0 \\
     gpt-4o-2024-05-13\cite{openaigpt4o}                    & N/A &	36.4 &	84.5 &	63.4 &	56.8 &	82.5 &	82.0 &	84.4 &	52.5 &	70.8 \\
     \hline
     Closed Source MLLM \\
     \hline
     gemini-1.5-pro(2024.05)\cite{google2024gemini15}       & N/A &	52.5 &	80.6 &	81.9 &	58.0 &	87.0 &	88.5 &	86.1 &	52.4 &	73.3 \\
     HyperCLOVA-X(2024.09)\cite{navercloud2024hyerclovax}   & N/A &	57.0 &	\textbf{93.3} &	79.1 &	44.8 &	83.5 &	88.1 &	86.1 &	55.7 &	74.0 \\
     claude-3-5-sonnet-20240620\cite{anthropic2024claude35} & N/A &	65.9 &	88.4 &	\textbf{93.3} &	67.4 &	94.0 &	93.3 &	90.7 &	62.8 &	80.6 \\
     gpt-4o-2024-05-13\cite{openaigpt4o}                    & N/A &	\textbf{62.5} &	89.2 &	\textbf{93.3} &	\textbf{69.5} &	\textbf{95.0} &	\textbf{95.4} &	\textbf{94.4} &	\textbf{66.1} &	\textbf{83.4} \\
    \bottomrule
  \end{tabular}
\end{adjustbox}
    \caption{\label{tab:main_results} \textbf{Results on various conventional benchmarks and KoNET.} These are achieved under the condition with CoT prompting and an off-the-shelf OCR API.}

\end{table*}

\subsection{Main Results}

Table~\ref{tab:main_results} outlines the main results, comparing KoNET performance with benchmarks like MathVista and ScienceQA. It also details subset performances for KoNET's components—elementary, middle, high school, and college exams.

Key insights include a general performance improvement with larger model sizes. Notably, there's a significant gap between closed-source APIs and open-source models, especially for KoNET, indicating open-source models lack tuning for Korean domains. Closed-source APIs likely excel due to Korea-targeted business strategies.

Models experience increased difficulty with advancing levels in the Korean curriculum, evident in subset performances. Complexity rises significantly at each educational stage, particularly in KoCSAT, highlighting the rigorous nature of these questions aligned with real-world standards.

The EXAONE-3.0-7.8B-Instruct model, a sovereign AI model specifically designed for the Korean language (bilingual in English and Korean), achieved a K-NET score of 45.5, significantly outperforming other models of similar size (7–8B). This suggests that benchmarks centered solely on English may not accurately assess AI performance in non-English or East Asian language environments. For instance, in the KoHGED (high school education exam), a question was based on the classic literary work Yongbieocheonga (Songs of the Dragons Flying to Heaven), a historical text from Korea's Joseon Dynasty published in 1445. This work is part of the standard curriculum in Korean education. Models lacking an understanding of the cultural context struggled to interpret the question and failed to provide the correct answer. In contrast, the EXAONE-3.0-7.8B-Instruct model successfully derived the correct response, demonstrating how linguistic and cultural specificity significantly impacts AI performance. Notably, open-source models such as EXAONE and Qwen2 have shown strong performance in Korean and East Asian contexts, highlighting the need for greater focus on non-English languages in future research and open-source AI development.

\subsection{Further Analyses}

\subsubsection*{Q1: Do MLLMs perform better on KoNET due to their support for multimodal inputs?}
Table~\ref{tab:main_results} indicates unexpected results, with MLLMs sometimes lagging behind LLMs on KoNET, contrary to other benchmarks. We analyze model pairs sharing LLM backbones in Table~\ref{tab:analysis_transition}. Without the off-the-shelf OCR assistance, closed-source MLLMs demonstrate competitive performance, comparable to LLMs with OCR support. However, many open-source MLLMs do not perform as effectively, revealing a specific challenge with text recognition in the Korean context.

\subsubsection*{Q2: Can CoT prompting improve performance on KoNET?}

As shown in Table~\ref{tab:analysis_transition}, CoT generally enhances performance across all models. Notably, this improvement is more pronounced in high-performing closed-source models compared to open-source models. This suggests that while CoT is beneficial, some open-source models are not yet fully optimized for reasoning in the Korean context, making CoT less effective.

\subsubsection*{Q3: Do AI models have similar error patterns to students?}

We compare human error rates on 327 questions with AI error rates. The human error rates in KoCSAT are derived from the Korean College Scholastic Ability Test (KoCSAT), which plays a crucial role in university admissions in South Korea. This exam is a large-scale standardized assessment taken by hundreds of thousands of students each year, who systematically prepare and sit for the test under controlled conditions. In this study, human error rates are calculated based on data from approximately 505K students, using official statistics published by the Korea Institute for Curriculum and Evaluation (KICE\footnote{\url{https://www.kice.re.kr}}). KICE is the official national institution responsible for the development and evaluation of all exams included in KoNET. 

To analyze error rates, we explore variability in model responses by assigning different personas~\citep{safdari2023personality} and adjusting parameters like temperature. Using gpt-4o-2024-05-13, the strongest of our test models, we create 10 personas,\footnote{Personas include `student,' `teacher,' `professor,' `engineer,' `scientist,' `mathematician,' `doctor,' `lawyer,' `master student,' and `PhD student.'} generating 10 responses per persona for a total of 120 responses. For gpt-4o-2024-05-13, gemini-1.5-pro, HyperCLOVA-X, and claude-3-5-sonnet-20240620, we use three personas (`student,' `teacher,' and `professor'),\footnote{Each persona undergoes 10 repeated experiments.} also generating 10 responses per persona for a total of 120 responses. This setup addresses the challenge of limited high-performing AI models by using personas to expand the response pool, thus enabling comprehensive trend comparisons between AI models and student groups.

\begin{figure}[t!]
  \centering
  \includegraphics[width=\linewidth]{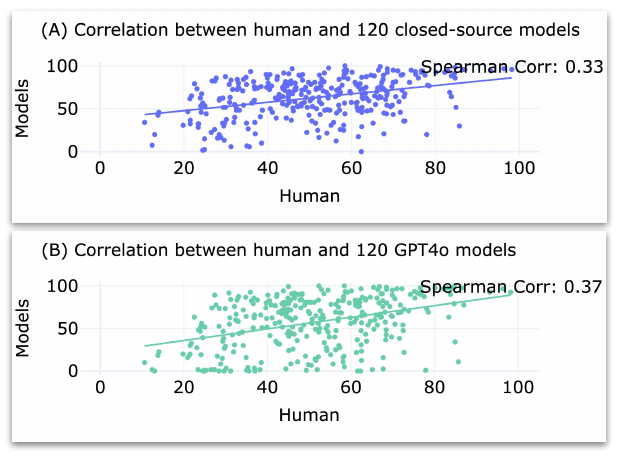}
\caption{\textbf{Correlation analysis of error rates.} The x-axis shows human error rates, and the y-axis displays error rates from closed-source models. Appendix~\ref{appendix:more_analyses_on_human_error_rate} offers a detailed discussion on the methods used to calculate these error rates.}
  \label{fig:error_rate}
\end{figure}

Figure~\ref{fig:error_rate} indicates a weaker than expected positive correlation. Detailed analysis shows AI models excel in comprehension tasks, likely due to human attention lapses, while humans perform better in memorization tasks, especially in long-tail questions for exams like the CSAT. These outcomes align with expectations and underscore the benchmark's value by integrating human error data, providing a rich resource for future studies.

\begin{table}[t!]
\begin{adjustbox}{width=\linewidth}
  \centering
  \begin{tabular}{lcccccc}
    \toprule
    \multirow{2}{*}{\textbf{Model}} & \multirow{2}{*}{\textbf{Size (B)}} & \multirow{2}{*}{\textbf{Mode}} & \multicolumn{2}{c}{\textbf{wo OCR}} & \multicolumn{2}{c}{\textbf{w OCR}} \\
    \cline{4-5}\cline{6-7}
    & & & \textbf{Direct} & \textbf{CoT} & \textbf{Direct} & \textbf{CoT} \\
    \hline
    \multirow{2}{*}{Qwen2-1.5B-Instruct} & \multirow{2}{*}{1.5} & Text & & & 14.7 & 19.2 \\
    & & Vision &  9.8 &	11.2 & 10.8 & 11.0 \\
    \hline
    \multirow{2}{*}{Phi-3.5-mini-instruct} & \multirow{2}{*}{3.8} & Text & & & 27.1 & 21.8 \\
    & & Vision & 21.1 &	4.4 & 24.9 & 10.9 \\
    \hline
    \multirow{2}{*}{Qwen2-7B-Instruct} & \multirow{2}{*}{7.0} & Text & & & 33.1 & 39.6 \\
    & & Vision & 21.9 &	33.9 & 35.7 & 34.3 \\
    \hline
    \multirow{2}{*}{Meta-Llama-3.1-70B-Instruct} & \multirow{2}{*}{70.0} & Text & & & 53.7 & 50.8 \\
    & & Vision & 22.1 & 4.2 & 45.5 & 30.7 \\
    \hline
    \multirow{2}{*}{gemini-1.5-pro} & \multirow{2}{*}{N/A} & Text & & & 64.3 & 66.4 \\
    & & Vision & 32.7 & 47.8 & 71.1 & 73.3 \\
    \hline
    \multirow{2}{*}{HyperCLOVA-X} & \multirow{2}{*}{N/A} & Text & & & 67.2 & 70.9 \\
    & & Vision & \textbf{69.5} & \textbf{75.2} & 69.5 & 74.0 \\
    \hline
    \multirow{2}{*}{claude-3-5-sonnet-20240620} & \multirow{2}{*}{N/A} & Text & & & 70.4 & 76.0 \\
    & & Vision & 40.2 & 73.5 & 71.1 & 80.6 \\
    \hline
    \multirow{2}{*}{gpt-4o-2024-05-13} & \multirow{2}{*}{N/A} & Text & & & 70.1 & 70.8 \\	
    & & Vision & 66.0 & 74.9 & \textbf{74.8} & \textbf{83.4} \\
    \bottomrule
  \end{tabular}
\end{adjustbox}
\caption{\label{tab:analysis_transition}
  \textbf{Comparison on common backbones.} This shows various LLMs with their corresponding MLLMs. 
}
\end{table}

\section{Conclusion}
\label{sec:conclusion}

We present KoNET as a benchmark for evaluating multimodal generative AI models using Korean educational tests. Our findings reveal varying performance with multimodal inputs and highlight specific challenges. The disparity between open and closed-source models points to the need for advancements in open-source models within non-English contexts. Our analysis of human error rates offers valuable insights into AI and human performance comparisons. Through KoNET, we aim to encourage research in multimodal and multilingual AI, thereby promoting inclusivity and diversity.

\section*{Limitations}

While KoNET serves as a valuable resource for assessing the intellectual capabilities of models through Korean educational tests, it does have certain limitations. Similar to many current benchmarks, KoNET primarily adheres to a multiple-choice QA format, which may not fully capture a model's capacity to articulate problem-solving processes. Although a small proportion of the questions are subjective (see Table~\ref{tab:compare_mcqa}), these generally involve short-response formats. To address this, future work could focus on evaluating models' reasoning abilities by incorporating rationales behind their answers. This advancement necessitates the development of comprehensive reference answers and a consideration of the increased computational costs involved.

Moreover, as is common with all benchmarks, periodic updates to the test set are necessary to mitigate potential biases and data contamination upon public release. Given that KoNET is based on annually updated national tests, it is inherently suited for regular renewal. We anticipate that our dataset construction methodology, along with the open-source dataset builder, will empower the research community to continuously update KoNET, ensuring its ongoing relevance and utility in advancing AI systems to better meet diverse needs.

\bibliography{anthology,custom}

\appendix

\section{Details on the KoNET Construction}
\label{appendix:examples_knet}

KoNET encompasses a wide range of subjects across each exam, as detailed in Table \ref{tab:subjects}. For K-GED (comprising KoEGED, KoMGED, KoHGED), core subjects are included as common components, while each exam features additional unique subjects. The KoCSAT comprises core subjects and optional subjects, with each optional subject further divided into specialized areas. Although students typically select specific subjects for their exams, this study includes questions from all subjects to ensure comprehensive coverage. All images within KoNET are presented in gray-scale, encapsulating the question, answer choices, and comprehension elements within a single image—a format that varies across problems. We adopt the simplest input method to evaluate both LLMs and MLLMs models. Each provided image is structured to contain both the question and all the information necessary to solve it. For text input, no additional text is provided beyond instruction-following prompts and OCR tokens (See Figure~\ref{fig:prompt}). This input format also allows us to indirectly assess the MLLMs models’ overall understanding of the image and their ability to recognize Korean characters.

\begin{figure*}[t!]
  \centering
  \includegraphics[width=0.95\linewidth]{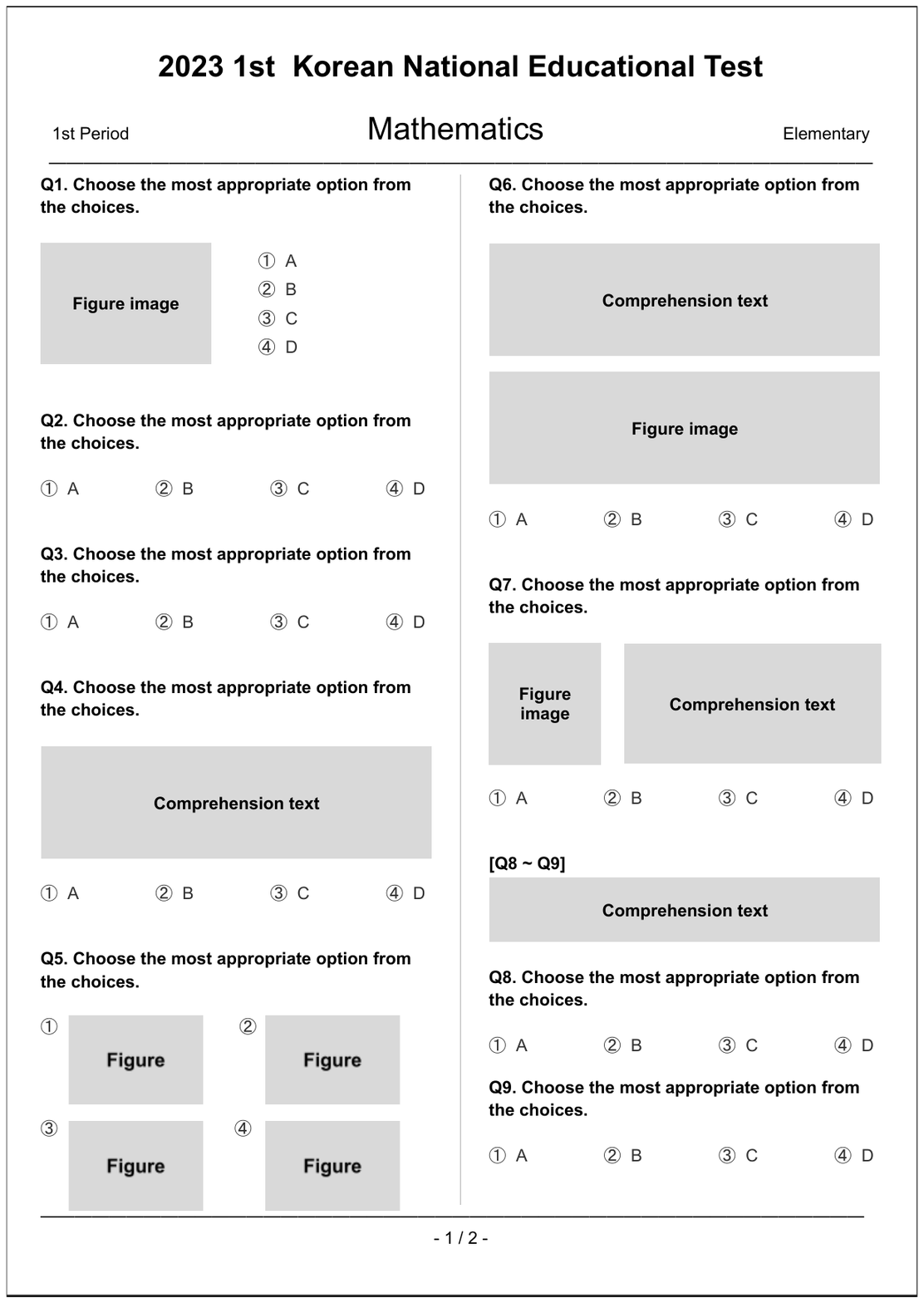}
\caption{\textbf{Illustrative Representation of the KoNET.} The test includes various types of questions, such as those requiring comprehension of images and queries, reading and understanding of lengthy texts, and simple knowledge-based queries.}
  \label{fig:example_knet}
\end{figure*}

KoNET is constructed by parsing publicly available official PDFs from the Korea Institute of Curriculum and Evaluation\footnote{\url{https://www.kice.re.kr}}. We remain mindful of licensing issues, acknowledging the inherent copyright of these questions. However, details regarding specific licensing terms remain elusive; the only guidance available from the Korea Institute of Curriculum and Evaluation indicates permission for non-commercial use. We uphold the copyrights of the original owners with utmost respect. Rather than distributing the data directly, we provide dataset builder code that allows users to convert downloaded official PDFs into benchmark-ready formats. In this paper, we include images that mimic various question types rather than actual problem images. The rendered images in the form of test sheets, based on these mimicked images, are shown in Figure \ref{fig:example_knet}. Actual problem images can be generated and reviewed using the provided dataset builder.

\begin{table}[t!]
\begin{adjustbox}{width=\linewidth}
  \centering
  \begin{tabular}{l|p{7cm}}
    \toprule
    \textbf{Test} & \textbf{Subjects} \\
    \midrule
    \textbf{KoEGED} & Korean, English, Mathematics, Social Studies, Science, Music, Physical Education, Ethics, Art, Practical \\
    \\
    \textbf{KoMGED} & Korean, English, Mathematics, Social Studies, Science, Music, Physical Education, Ethics, Art, Information, Technology \\
    \\
    \textbf{KoHGED} & Korean, English, Mathematics, Social Studies, Science, Music, Physical Education, Ethics, Art, Technology, Korean History \\
    \\
    \textbf{KoCSAT} & Korean (Common), Korean (Speech Writing), Korean (Language and Media), Mathematics (Common), Mathematics (Statistics), Mathematics (Calculus), Mathematics (Geometry), English, Korean History, Social Studies (Every Ethics), Social Studies (Ethical Ideology), Social Studies (Korean Geography), Social Studies (International Geography), Social Studies (East Asia History), Social Studies (International History), Social Studies (Economics), Social Studies(Politics and Law), Social Studies(Social Culture), Science (Physics I), Science (Chemistry I), Science (Bio Science I), Science (Earth Science I), Science (Physics II), Science (Chemistry II), Science (Bio Science II), Science (Earth Science II), Job Studies (Successful Career Life), Job Studies (Agricultural Technology), Job Studies (General Industry), Job Studies (Commercial Economy), Job Studies (Fisheries Shipping Industry), Job Studies (Human Development), Second Language (German), Second Language (French), Second Language (Spanish), Second Language (Chinese), Second Language (Japanese), Second Language (Russian), Second Language (Arabic), Second Language (Vietnamese), Second Language (Chinese characters) \\
    \bottomrule
  \end{tabular}
\end{adjustbox}
  \caption{\label{tab:subjects}
    \textbf{List of subjects categorized under various Korean educational tests.} KoEGED represents subjects for elementary-level general education (10 subjects), KoMGED covers middle-level general education (11 subjects), and KoHGED encompasses high school-level general education (11 subjects). KoCSAT includes the 41 subjects evaluated in the Korean College Scholastic Ability Test, spanning multiple disciplines, including languages, mathematics, sciences, social studies, and job studies.
  }
\end{table}

\section{Details of the Used Prompts}
\label{appendix:prompts}

\begin{figure*}[t!]
  \centering
  \includegraphics[width=0.9\linewidth]{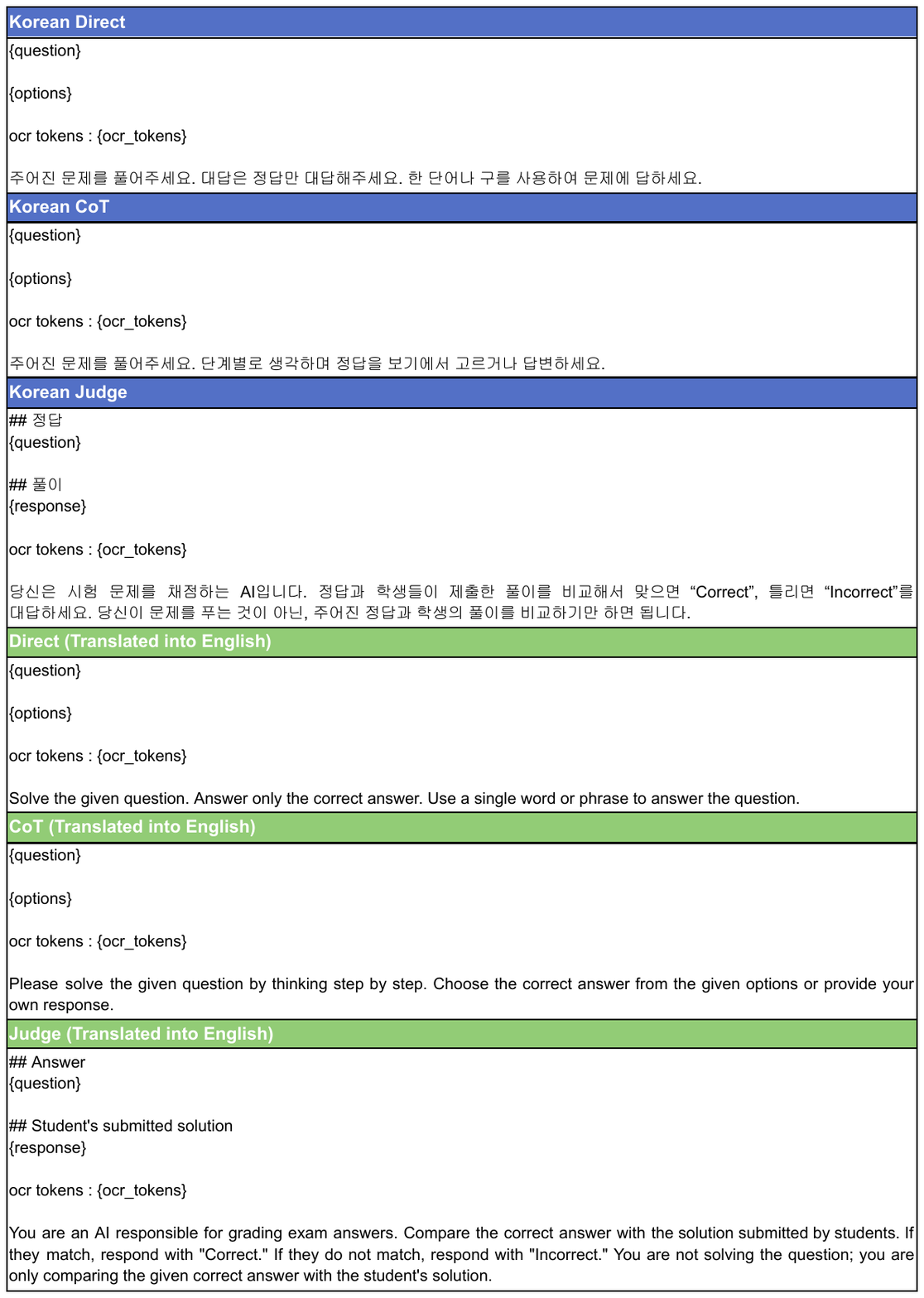}
    \caption{\textbf{Examples of prompt formats used in the study.} These include Direct prompts for answer extraction, CoT (Chain-of-Thought) prompts for reasoning-based inference, and Judge prompts for evaluating the accuracy of generated responses.}
  \label{fig:prompt}
\end{figure*}

In this study, we use Korean prompts to generate and assess the response generation capabilities of the models. Two types of prompts are employed: the Direct prompt and the Chain of Thought (CoT) prompt. The Direct prompt involves extracting answers directly from the provided options for each question. Conversely, the CoT prompt allows the model to reason through the problem to infer the answer. Additionally, a Judge prompt is used within the CoT framework to evaluate the responses generated by comparing them with the correct answers. While the original prompts are in Korean, English translations are also provided for reference. The format of these prompts is exemplified in Figure \ref{fig:prompt}.

\begin{figure}[t!]
  \centering
  \includegraphics[width=\linewidth]{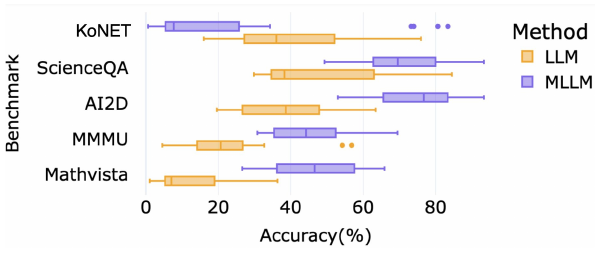}
  \caption{\textbf{Performance of LLMs and MLLMs across Previous benchmarks and KoNET.} These present a performance comparison between LLMs and MLLMs across various benchmarks, including KoNET. These illustrate the accuracy distribution for each model type, but KoNET shows a different distribution trend between LLMs and MLLMs compared to other benchmarks.}
  \label{fig:comparision_public_benchamark}
\end{figure}

\section{Additional Analysis}
\label{appendix:additional_analysis}

\subsection{On the Performance Gap Between LLMs and MLLMs}

Figure \ref{fig:comparision_public_benchamark} illustrates the score distribution of LLMs and MLLMs on both conventional benchmarks and KoNET. As shown in our work, the KoNET reveals a distinct distribution pattern compared to traditional benchmarks. Notably, MLLMs underperform relative to LLMs. As analyzed in the paper, we suggest that public LLMs may actually achieve better performance when supported by Korean OCR and many commercially available MLLMs are less effective in processing non-English contexts. This finding provides a novel perspective for model analysis that diverges from traditional benchmarks.

\subsection{Comparison of LLM-as-a-Judge with Manual Grading}

To see whether LLM-as-a-Judge provide similar user experience or performance to manual grading, we conduct an additional analysis on this. Given the multiple-choice nature of the tests and the potential for varying text responses, we adopt the LLM-as-a-Judge strategy to ensure grading accuracy. Table~\ref{tab:judge} indicates that this approach closely mirrors manual grading results, demonstrating its reliability and potential as an efficient evaluation method.

\begin{table}[t!]
\begin{adjustbox}{width=\linewidth}
  \centering
  \begin{tabular}{lccccc}
    \toprule
      & \textbf{Test 1} & \textbf{Test 2} & \textbf{Test 3} & \textbf{Test 4} & \textbf{Test 5} \\
    \hline
    \textbf{Accuracy} & 96.9\% & 98.3\% & 98.2\% & 97.4\% & 98.2\% \\
    \bottomrule
  \end{tabular}
\end{adjustbox}
\caption{\label{tab:judge} \textbf{Agreement Rate Between Human Evaluation and Judge Model.} When using the LLM-as-a-Judge approach, results may vary slightly with each evaluation. To ensure consistency, we conduct evaluations five times to assess whether the LLM-as-a-Judge method aligns closely with answers annotated manually by the authors. When considering the authors' evaluation results as the ground truth, we find that the accuracy is consistently high. This suggests that LLMs can reliably substitute human evaluators with a high degree of confidence.}
\end{table}

\subsection{Analysis of Human Error Rates}
\label{appendix:more_analyses_on_human_error_rate}

\begin{figure*}[t!]
  \centering
  \includegraphics[width=\linewidth]{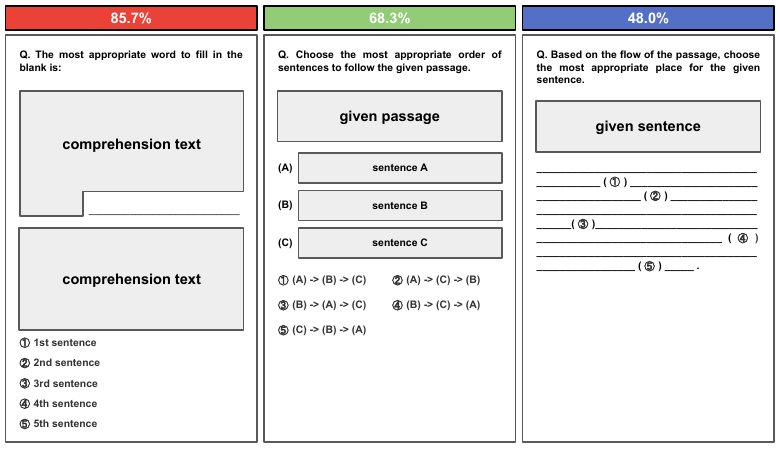}
  \caption{\textbf{Examples of human error rate.} These illustrates human error rates across three types of comprehension tasks: sentence selection (left), sentence ordering (middle), and sentence insertion (right). The percentages at the top represent the error rates calculated based on responses from students. Higher error rates indicate more challenging tasks requiring deeper comprehension. Notably, as the complexity of the comprehension text increases, the error rate also rises, suggesting a greater cognitive load in understanding and structuring the given information.}
  \label{fig:example_error_rate}
\end{figure*}

We employ the error rates from the KoCSAT to assess and compare the performance of models against human performance. Human error rates range from 10.6\% to 98.2\%, as illustrated in Figure~\ref{fig:example_error_rate}. 

In the first analysis, we calculate model error rates using four closed-source MLLM APIs. For each model, we configure ten personas (i.e., different system messages), set the temperature to 1.0, and generate outputs three times. 

\begin{figure*}[t!]
  \centering
  \includegraphics[width=\linewidth]{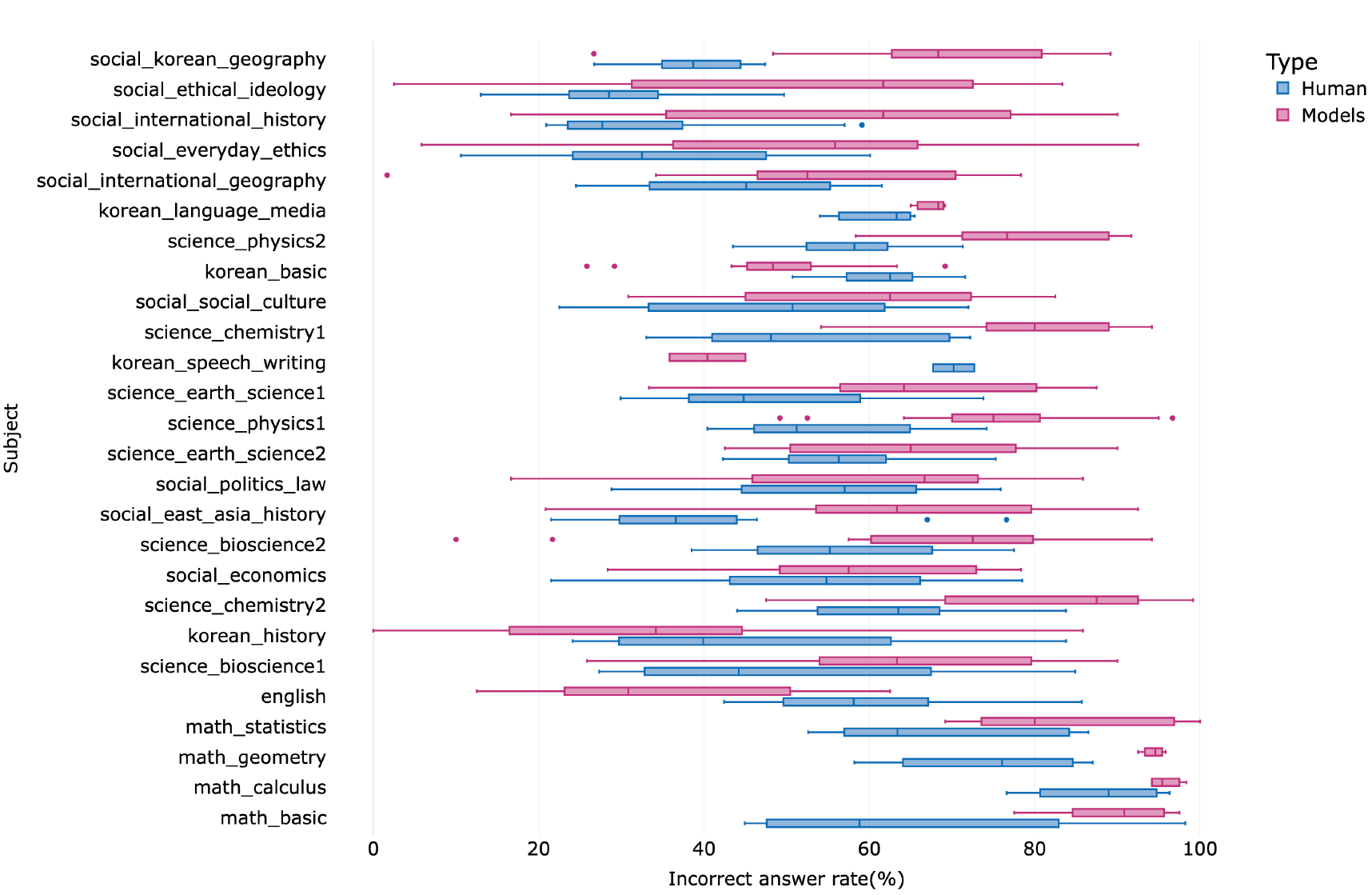}
  \caption{\textbf{Distribution of human and models error rate by subjects.} These compares the error rate distributions between humans (blue) and models (pink) across various academic subjects. The x-axis represents the error rate, while the y-axis lists different subjects, covering social sciences, natural sciences, Korean language, history, and mathematics. The varying distributions highlight the differences in performance between humans and models, with some subjects showing a greater disparity.}
  \label{fig:error_rate_subjects}
\end{figure*}

\begin{figure*}[t!]
  \centering
  \includegraphics[width=\linewidth]{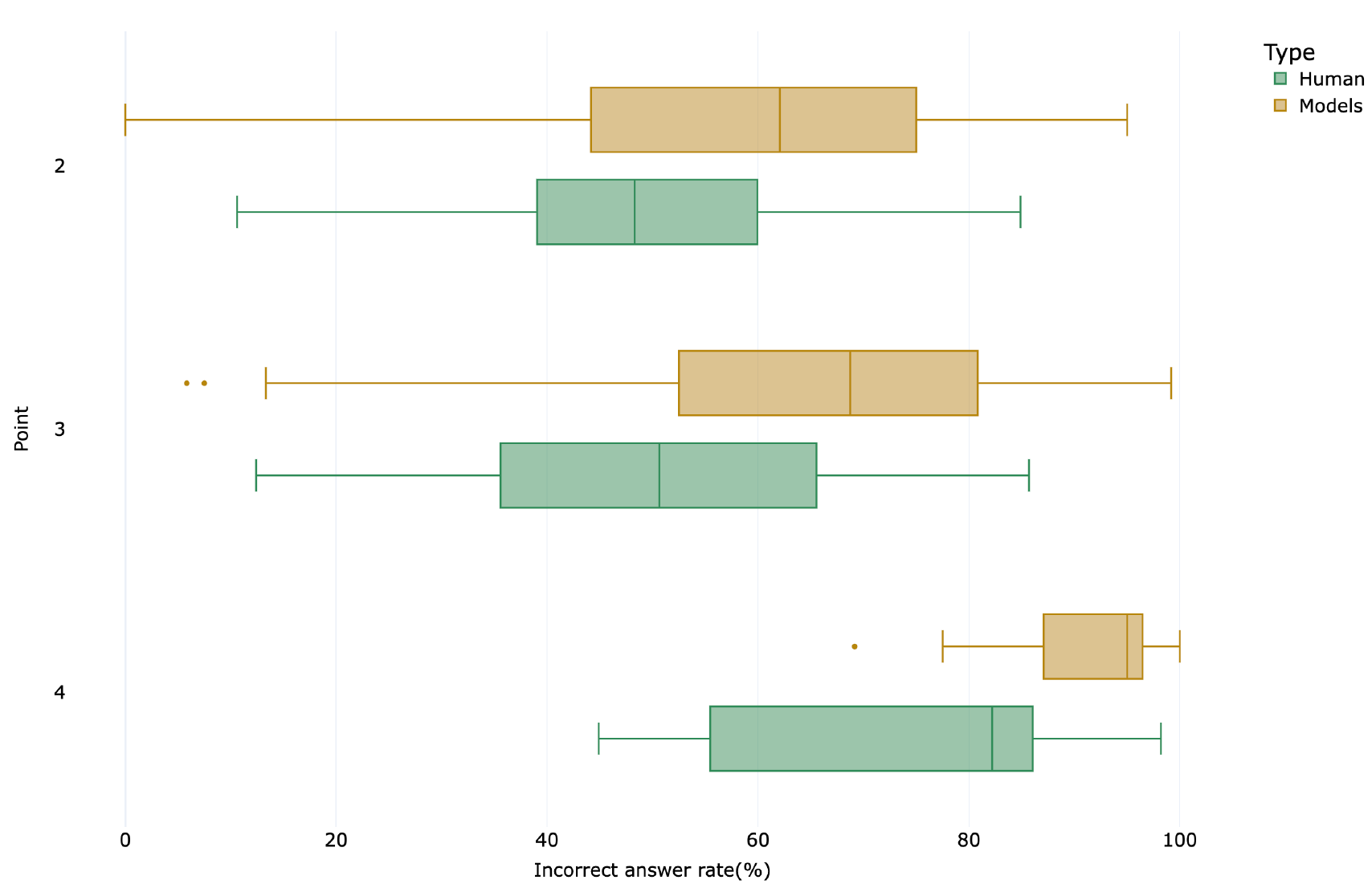}
  \caption{\textbf{Distribution of human and models error rate by points.} These presents the error rate distribution of humans (green) and models (brown) based on different point values assigned to questions. The x-axis represents the percentage of incorrect answers, while the y-axis categorizes questions by their point values. Higher-point questions generally require deeper reasoning and comprehension, which is reflected in the increasing error rates for both humans and models.}
  \label{fig:error_rate_points}
\end{figure*}

In the second analysis, we utilize the GPT-4o model across ten personas, generating twelve distinct responses per persona. We then compute the model error rates and compare them with the human error rates. Figure~\ref{fig:error_rate_subjects} illustrates the distribution of error rates across subjects, while Figure~\ref{fig:error_rate_points} provides a point-by-point comparison of human and model error rates.

This rigorous analysis enhances our understanding of model performance relative to human benchmarks, offering valuable insights into the strengths and limitations of current MLLMs in processing complex educational content.

\subsection{Multilingual Ability Assessment}

\begin{figure*}[t!]
  \centering
  \includegraphics[width=\linewidth]{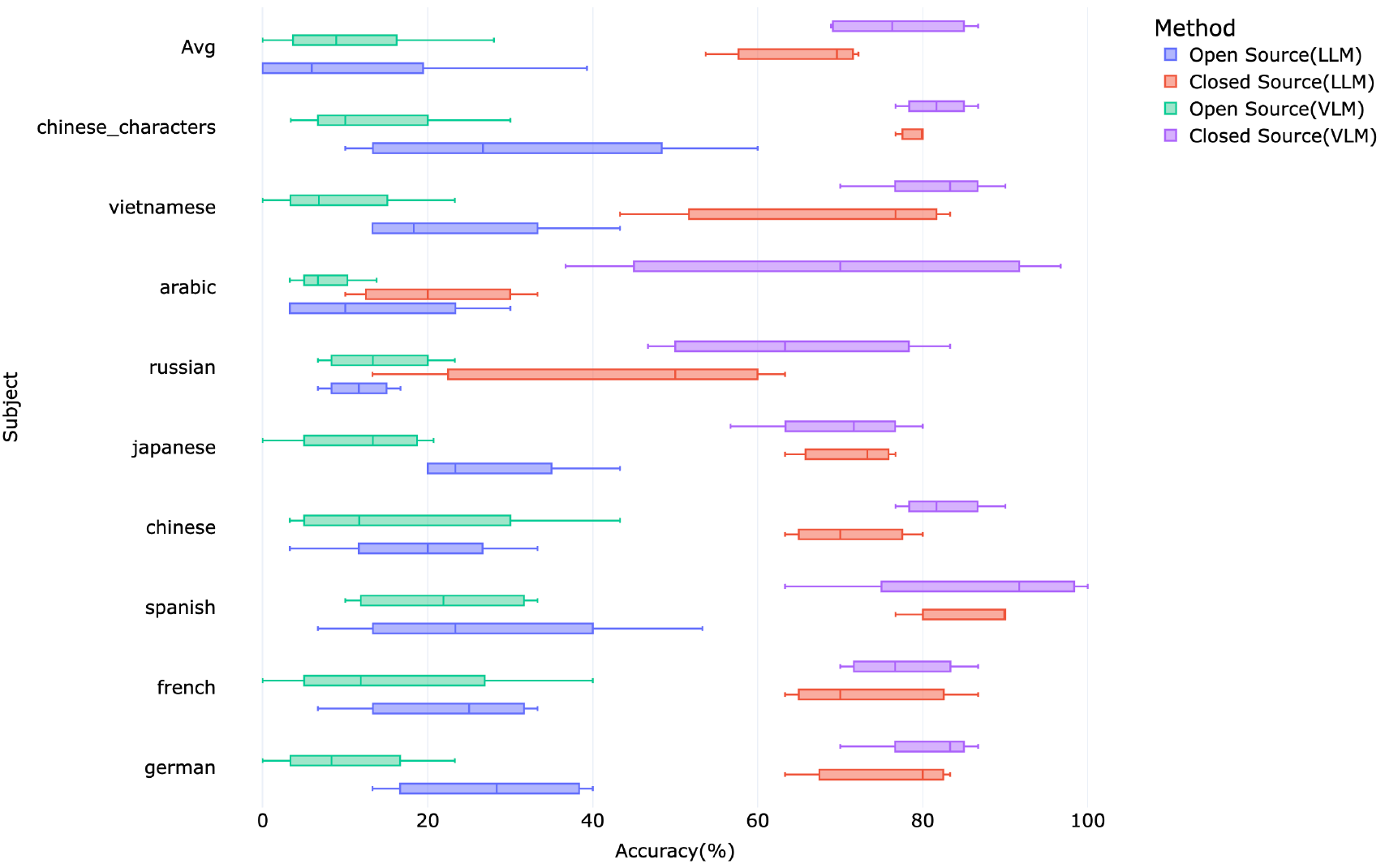}
  \caption{\textbf{Performance of multilingual ability.} These illustrations depict the accuracy distribution of various models across multiple languages, highlighting their multilingual capabilities. The x-axis represents accuracy percentages, while the y-axis lists different languages. In general, Open Source models tend to support a narrower range of languages fluently compared to Closed Source models. However, even among Closed Source LLMs, performance tends to decline for certain languages; for instance, Arabic differs from English in writing direction, which can impact model performance.}
  \label{fig:multilingual}
\end{figure*}

We assess multilingual capabilities using specific subjects from KoNET. The KoCSAT includes subjects for nine different languages. Traditionally, multilingual capabilities are evaluated by translating English-based benchmarks into other languages or by making indirect comparisons using benchmarks crafted in different linguistic regions. However, the multilingual subjects in KoCSAT consist of independent questions with comparable difficulty levels, enabling a more equitable and valid comparison of multilingual abilities. Figure \ref{fig:multilingual} illustrates the multilingual capabilities across different model types.

\end{document}